\documentclass[letterpaper, 10 pt, conference]{ieeeconf}  

\IEEEoverridecommandlockouts                              

\usepackage{graphics} 
\usepackage{epsfig} 
\usepackage{mathptmx} 
\usepackage{times} 
\usepackage{amsmath} 
\usepackage{amssymb}  

\usepackage{cite}
\usepackage[separate-uncertainty=true]{siunitx}
\usepackage{comment}
\usepackage{kotex}

\title{\LARGE \bf
Polymander II: an amphibious salamander-inspired robot with contact and flow sensors
}

\author{Qiyuan Fu$^{\dagger,1}$, Sudong Lee$^{\dagger,2}$, Andrea Grillo$^{1}$, Jonathan Arreguit$^{1,3}$, Louis Gevers$^{1}$,\\Josie Hughes$^{2}$, and Auke J. Ijspeert$^{1}$
\thanks{$\dagger$ These authors equally contributed to this work.}
\thanks{* This work was supported by European Research Council (Grant Agreement No. 951477) }
\thanks{$^{1}$Q. Fu, A. Grillo, J. Arreguit, L. Gevers, and A. J. Ijspeert are with the Biorobotics Laboratory, EPFL, 1015 Lausanne, Switzerland.
        (email: qiyuan.fu@epfl.ch; andrea.grillo@epfl.ch; jonathan.arreguitoneill@epfl.ch; louis.gevers@epfl.ch)}
\thanks{$^{2}$S. Lee and J. Hughes are with the CREATE Lab, EPFL, 1015 Lausanne, Switzerland.
        (email: sudong.lee@epfl.ch; josie.hughes@epfl.ch)}
\thanks{$^{3}$J. Arreguit is with Innobridge Services Sàrl, 1015 Lausanne, Switzerland.
        (email: j.arreguitoneill@innobridge.com)}
\thanks{© 2026 IEEE.  Personal use of this material is permitted.  Permission from IEEE must be obtained for all other uses, in any current or future media, including reprinting/republishing this material for advertising or promotional purposes, creating new collective works, for resale or redistribution to servers or lists, or reuse of any copyrighted component of this work in other works. This work has been accepted for publication in the 2026 \it{International Conference on Robotics and Automation (ICRA)}, Vienna, Austria.}
}

\begin{document}

\maketitle
\thispagestyle{empty}
\pagestyle{empty}

\begin{abstract}
 Robots benefit from sensory information to coordinate body movement, gain robustness against perturbations, and transition between different modes to adapt to various terrains. However, few amphibious robots can sense interactions with both terrestrial and aquatic environments. In this paper, we present a solution that uses Hall-effect sensors to sense foot contact forces and lateral hydrodynamic forces on a salamander-inspired amphibious robot.
 With two bus lines, the robot can simultaneously acquire this exteroceptive information at more than 500~Hz and proprioceptive information, such as joint positions and loads, at 100~Hz. The Hall-effect sensors used are compact, making them suitable for embedding in multiple positions within a robot, and exhibit high sensitivity to small forces. Moreover, because the sensor can be positioned separately from the measured object, waterproofing can be implemented with relative ease. Our tests demonstrate the robot's capabilities in traversing amphibious environments and its potential in using feedback control for more complex locomotion tasks.
\end{abstract}

\section{Introduction}
Amphibious robots could enable autonomous operation in challenging environments which are typically difficult for humans or more traditional field robots to access~\cite{delmericoCurrentStateFuture2019}. The mixture of water, flowable media, and solid ground with obstacles poses significant physical challenges to the robot's structure and requires multimodal control to traverse the varied environment. Amphibious animals such as salamanders~\cite{ijspeertAmphibiousSprawlingLocomotion2020} demonstrate versatile multimodal behaviors in such environments, including walking, swimming, and smooth transitions between these modes, inspiring the control of amphibious robots.

Inspired by their biological counterparts, many amphibious salamander robots have been developed as platforms for studying their locomotion control, and also to be used for search and rescue tasks~\cite{ijspeertAmphibiousSprawlingLocomotion2020}. These robots mimic the salamander's flexible spine, which has been shown to increase the robot's efficiency for both swimming and terrestrial locomotion~\cite{karakasiliotisWhereAreWe2013}. Biologically inspired controllers such as central pattern generators (CPGs) have been developed, which enable multimodal locomotion of the complex system with limited inputs~\cite{ryczkoWalkingSalamandersMolecules2020}. Modeling of such controllers has demonstrated the potential of using sensory feedback from interacting with the environment to coordinate the body movement, such as using foot contact feedback to coordinate leg movement in walking \cite{suzuki2021SprawlingQuadrupedRobot, arreguit2025farms} and using hydrodynamic force feedback to coordinate spine bending in swimming or crawling \cite{thandiackalEmergenceRobustSelforganized2021, yasui2025MultisensoryFeedbackMakes}. Aside from coordination, sensory feedback can also help the robot gain robustness against perturbations to signal transmissions and adapt to varying environments. However, most of the previous salamander robots~\cite{ijspeertAmphibiousSprawlingLocomotion2020} lack the ability to sense such physical interactions with the environment. Whilst a few salamander robots have contact force sensors~\cite{leung2025BioInspiredPlasticNeural, meloAnimalRobotsAfrican2023,suzuki2021SprawlingQuadrupedRobot}, these are typically not used in amphibious environments. This has greatly hindered the verification of hypothesized force-feedback amphibious locomotion controllers in real-world environments ~\cite{arreguit2025farms}.

\begin{figure}
    \centering
    \includegraphics[width=1.0\linewidth]{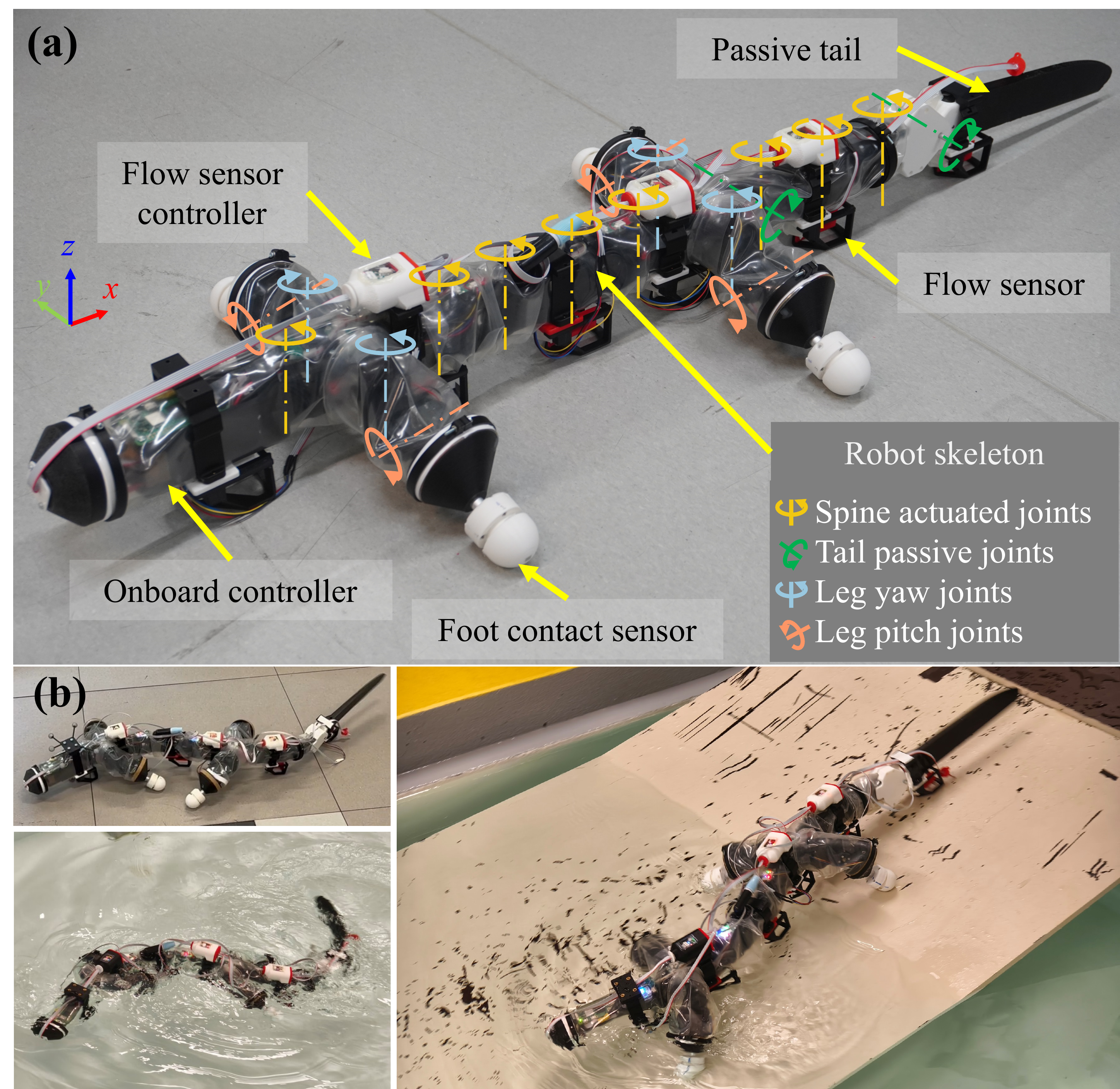}
    \caption{Overview of Polymander II. (a) Major components. (b) Applications in amphibious environments.
    }
    \label{fig:fig_overall}
\end{figure}

The lack of sensing is largely due to the challenges of installing waterproof force sensors~\cite{renResearchStatusBionic2021}. Off-the-shelf load cells are often bulky and heavy to incorporate in amphibious robots with multiple degrees of freedom (DoFs), which often requires the robot to scale up~\cite{yasui2025MultisensoryFeedbackMakes}. The rigid structures are also vulnerable to overloads caused by impacts with the ground or obstacles. In contrast, force sensors based on small and lightweight Hall-effect sensors that detect displacement of magnets attached to custom elastic structures impose minimal design and volume constraints on robots, allowing their integration without significantly affecting robot dynamics or workspace~\cite{Lu2022GTac, Signor2023Mass}. Furthermore, because the Hall-effect sensor and the target magnet do not require physical contact, the sensor electronics can be embedded within the robot while installing the elastic structures externally, drastically reducing the difficulty of waterproofing and the risk of damaging the entire sensor.

Another major challenge of designing sensorized amphibious salamander robots is waterproofing the flexible spine and the legs. Limited choices are available for compact waterproof actuators that provide rich feedback~\cite{kimDevelopmentBioinspiredMultimodal2024, quAmphibiousRoboticDog2025, panMiniatureDeepseaMorphable2025}. Waterproof rigid shells~\cite{crespiSalamandraRoboticaII2013, liljebackMambaWaterproofSnake2014,hirose2009SnakelikeRobotsTutorial, fernandezAquaMILRDesignUntethered2024, prahacsLeggedAmphibiousMobile2011} require a large volume around rotational joints. Tailored flexible dry suits~\cite{karakasiliotis2016CineradiographyBiorobotsApproach, meloAnimalRobotsAfrican2023,kelasidiExperimentalInvestigationEfficient2015, anastasiadisEellikeRobotSwims2024, yasui2025MultisensoryFeedbackMakes} can minimize the overall size while allowing a large range of motion, but they are costly to produce and difficult to inspect for leakage or control the buoyancy.

In this work, we develop miniature Hall-effect sensors to detect contact and hydrodynamic forces in amphibious robots. We use identical electronics for both types of sensors, but with different elastic structures that deform under different loads. This allows easier sealing inside custom-shaped housings and sharing of components for maintainability. Custom serial bus lines are used to manage multiple peripheral sensors aside from the actuators, leaving space for future addition or removal of devices. To test the sensors, we develop Polymander II, an amphibious salamander-inspired robot modified from \cite{geversInvestigatingEffectMorphology2025}. We collect force sensor readings at high speeds while controlling joint angles using CPGs during walking and swimming. The force signals match the locomotion patterns and can be used to trigger a smooth transition from walking to swimming. These demonstrate that our sensors and robot are promising for future experiments of multimodal sensory feedback controllers in amphibious environments and applications that require traversing such environments.

\section{Sensors} \label{section_sensor}
To sense foot contacts and lateral hydrodynamic forces that benefit amphibious locomotion~\cite{suzuki2021SprawlingQuadrupedRobot,arreguit2025farms,thandiackalEmergenceRobustSelforganized2021,yasui2025MultisensoryFeedbackMakes}, we installed contact sensors inside each foot and six flow sensors under the spine.

\subsection{Foot contact sensor}

\subsubsection{Design}
\begin{figure}
    \centering
    \includegraphics[width=1.0\linewidth]{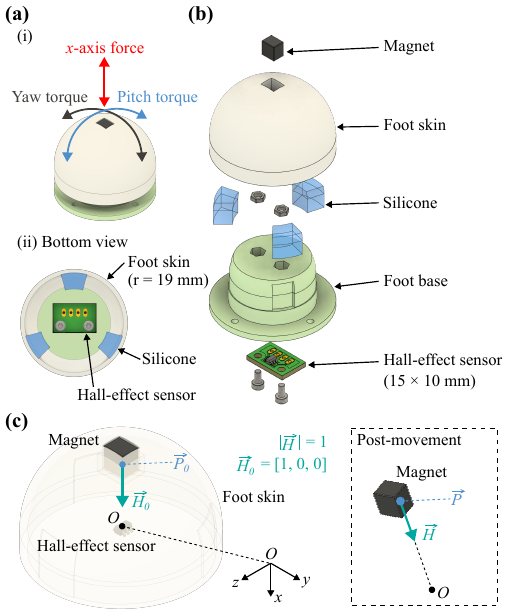}
    \caption{Design of the foot contact sensor.
    (a) (i) Assembled design and sensor degrees of freedom, and (ii) bottom view.
    (b) Expansion view.
    (c) Configuration of the foot contact sensor, including the magnet location ($\vec{P}$), the magnetic field direction ($\vec{H}$), and the origin ($O$) in both the initial and post-movement states.
    }
    \label{fig:FootSensor_01}
\end{figure}
To measure the contact forces against 3D terrain, the foot contact sensor measures three types of torques or forces (Fig.~\ref{fig:FootSensor_01}(a)(i)). To facilitate this estimation, the sensor provides three degrees of freedom: rotation about the $y$-axis (pitch), rotation about the $z$-axis (yaw), and translation along the $x$-axis. The Hall-effect sensor is embedded at the center of the foot base (Fig.~\ref{fig:FootSensor_01}(a)(ii)). Three silicone structures are attached to the foot base (Fig.~\ref{fig:FootSensor_01}(b)), on which the foot skin (radius = 19 mm) is mounted. Owing to the flexibility of silicone, the foot skin can move relative to the foot base, enabling motion with three DoFs. Fabricated by the relatively stiff material (Dragon Skin 20, Smooth-On), the silicone structure offers limited compression while allowing shear and bending. Because the Hall-effect sensor is positioned at the center of the foot skin, its location coincides with the rotation center.
Therefore, in modeling, the origin of the local frame of the sensor system can be assigned to the Hall-effect sensor, which simplifies the calculation of the position vector.
The sensor detects changes in the magnetic flux generated by the magnet attached to the foot skin. Since the three-axis magnetic flux is not independently related to the measuring loads along each axis, a model is required to map the flux to forces.

\subsubsection{Analytical Modeling} \label{sec:FootSensor_Model}
The 6 DoF movement of the magnet is constrained to 3 DoFs, translation along the $x$-axis and rotation around the $y$- and $z$-axes, because of the low compressibility of the rigid silicone. These 3 DoFs correspond one-to-one with the magnet location of $x$-, $y$-, and $z$- axes. We estimate the location of the magnet through the model using 3-axis magnetic flux, and this location is mapped to the applied force or torques.
Ultimately, our objective is to measure the $x$-axis normal force, pitch, and yaw torques by using 3-axis magnetic flux information.

In Fig.~\ref{fig:FootSensor_01}(c), the magnetic flux is as follows:
\begin{equation}
\vec{B} = \left( \mu_{r}\mu_{0}M_{T} / 4\pi \right) \left[ 3\left(\vec{H}\cdot\vec{P}\right)\vec{P} - \vert\vec{P}\vert^{2} \vec{H} \right] / \vert\vec{P}\vert^{5}
\label{eqn:magnetic_flux_density_1}
\end{equation}
where $\vec{P}$ is the location of the magnet ($\left[p_{x}, p_{y}, p_{z}\right]$), $\vec{H}$ is the magnetic field direction ($\left[h_{x}, h_{y}, h_{z}\right]$) of the magnet, $M_{T}$ is the magnetic moment, and $\mu_{r}$ and $\mu_{0}$ are the relative permeability of the medium and the air magnetic permeability, respectively~\cite{Zhang2023Analytical, Hu2005Efficient, Field2014Cheng}.
Since $\mu_{r}\mu_{0}M_{T}/4\pi$ is simplified to the constant $N_{T}$, and $\vec{H}$ can be expressed as $-\vec{P}/\vert\vec{P}\vert$, (\ref{eqn:magnetic_flux_density_1}) becomes as follows:
\begin{equation}
\vec{B} = -N_{T} \left[ 2\vec{P} / \vert\vec{P}\vert^{4} \right]
\label{eqn:magnetic_flux_density_2}
\end{equation}
By measuring $\vec{B}$ with a Hall-effect sensor and substituting it into (\ref{eqn:magnetic_flux_density_2}), $\vec{P}$ is determined.

\subsubsection{Characterization}
\begin{figure}
    \centering
    \includegraphics[width=1.0\linewidth]{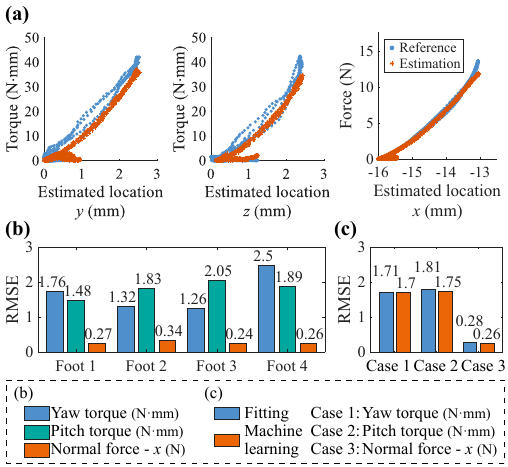}
    \caption{Characterization results.
    (a) Mapping results of applied forces and estimated magnet location using the second-order fitting method.
    (b) Estimation accuracy (root mean square errors) for the four individual feet.
    (c) Comparison of estimation accuracy between the fitting method and the machine learning method.
    }
    \label{fig:FootSensor_03}
\end{figure}
The location of the magnet, estimated using the model presented in Section~\ref{sec:FootSensor_Model}, is mapped to the measured forces through both fitting and machine learning (ML) approaches. The magnetic flux collected by the Hall-effect sensor is post-processed using a low-pass filter with a cut-off frequency of \SI{3.6}{\hertz} to remove noise. The filtered magnetic flux ($\vec{B}$) is then substituted into (\ref{eqn:magnetic_flux_density_2}) to solve for $\vec{P}$. The three components of $\vec{P}$ ($p_{x}$, $p_{y}$, $p_{z}$) serve as inputs for mapping pitch torque, yaw torque, and $x$-axis force. In the fitting method, a separate model is fitted to each type of force using second-order polynomials. In the ML method, a single model is employed to estimate all three forces simultaneously from the three input components. Reference force values are obtained using a load cell, while reference torques are calculated by multiplying the measured force by the distance from the rotational center.

Characterization is performed for each of the four feet using a load cell and a sensor-connecting jig. The training data are collected over 10 cycles for each foot and load type, while evaluation data are obtained from two independent cycles per foot. The fitting results of the characterization for the estimated magnet positions are shown in Figs.~\ref{fig:FootSensor_03}(a), (b).
For torque, the root mean square error (RMSE) ranges from \SI{1.26}{\newton\cdot\milli\meter} to \SI{2.5}{\newton\cdot\milli\meter}, with an average RMSE of \SI{1.76}{\newton\cdot\milli\meter}. For force, the RMSE ranges from \SI{0.24}{\newton} to \SI{0.34}{\newton}, with an average RMSE of \SI{0.28}{\newton}.
For the machine learning method\footnote{The machine learning uses a deep neural network (DNN) model with four hidden layers (128, 256, 128, and 64 nodes) based on PyTorch ([Online]. Available: https://pytorch.org/). The model is optimized using the Adam optimizer with a learning rate of 0.001 and trained for 1024 epochs with a batch size of 64.}, the average RMSE values for torque and force are \SI{1.73}{\newton\cdot\milli\meter} and \SI{0.28}{\newton}, respectively. As shown in Fig.~\ref{fig:FootSensor_03}(c), although the use of ML improved accuracy, the enhancement is modest. Considering the future integration of sensor processing algorithms within the robot, the fitting method is selected due to its lower computational cost.

\begin{figure}
    \centering
    \includegraphics[width=1.0\linewidth]{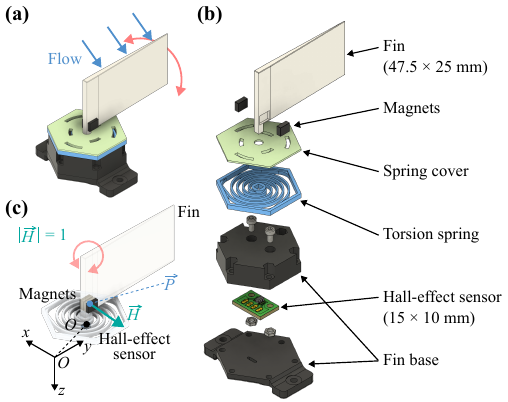}
    \caption{Design of the flow sensor.
    (a) Assembled design and sensor degrees of freedom.
    (b) Expansion view.
    (c) Configuration of the flow sensor, including the magnet location ($\vec{P}$), the magnetic field direction ($\vec{H}$), and the origin ($O$).
    }
    \label{fig:FinSensor_01}
\end{figure}

\subsection{Flow sensor}

\subsubsection{Design}
The flow sensor rotates about the $z$-axis in response to water flow (Fig.~\ref{fig:FinSensor_01}(a)). A fin (25 $\times$ 47.5 mm) is connected to a torsion spring (Fig.~\ref{fig:FinSensor_01}(b)), which allows it to rotate under the applied flow and return to its original position when no force is present. All degrees of freedom other than rotation about the $z$-axis are constrained by the spring cover and fin base. Magnets are positioned at the bottom of the fin, generating changes in magnetic flux due to the fin's rotation. This flux variation is measured by a Hall-effect sensor embedded in the fin base. The six flow sensors were clamped under the first, second, fourth, sixth, and eighth spine links as well as the passive tail and protected by 3-D printed cages against ground contact without blocking the water flow.

\subsubsection{Analytical Modeling} \label{sec:FinSensor_Model}
The fin has only 1 DoF ($z$-axis rotation), as shown in Fig.~\ref{fig:FinSensor_01}. This motion alters the $x$- and $y$-axes components of the magnetic flux direction and simultaneously changes the $x$- and $y$-axes location of the magnet. 
In this configuration, $z$-axis location of the magnet ($p_{z}$) is constant ($d_{z0}$) and the $z$-axis value of the magnetic flux direction remains zero. Since the magnitude of the magnetic flux direction ($\vec{H}$) is one, the magnitude of the $h_{x}$ is expressed as $\sqrt{1-h_{y}^{2}}$.

Fig.~\ref{fig:FinSensor_01}(c) shows the configuration of the fin sensor.
Applying the given conditions to (\ref{eqn:magnetic_flux_density_1}), the following three equations are derived:
\begin{equation}
    \begin{aligned}
        &B_{x} = N_{T} \left[ 3\left(p_{x}^{2}\sqrt{1-h_{y}^{2}}+p_{x}p_{y}h_{y}\right) - \vert\vec{P}\vert^{2}\sqrt{1-h_{y}^{2}} \right] / \vert\vec{P}\vert^{5}\text{,} \\  
        &B_{y} = N_{T} \left[ 3\left(p_{x}p_{y}\sqrt{1-h_{y}^{2}}+p_{y}^{2}h_{y}\right) - \vert\vec{P}\vert^{2}h_{y} \right] / \vert\vec{P}\vert^{5}\text{,} \\
        &\text{and}~B_{z} = N_{T} \left[ 3d_{z0}\left(p_{x}\sqrt{1-h_{y}^{2}}+p_{y}h_{y}\right) \right] / \vert\vec{P}\vert^{5}\text{.}
    \end{aligned} 
    \label{eqn:magnetic_flux_density_3}
\end{equation}
By substituting the measured magnetic flux from the Hall-effect sensor into the three equations (\ref{eqn:magnetic_flux_density_3}), $p_{x}$, $p_{y}$, and $h_{y}$ can be determined.
Finally, the location of the magnet is not only converted into a rotational angle but also utilized for mapping water flow forces.

\subsubsection{Characterization}
\begin{figure}
    \centering
    \includegraphics[width=1.0\linewidth]{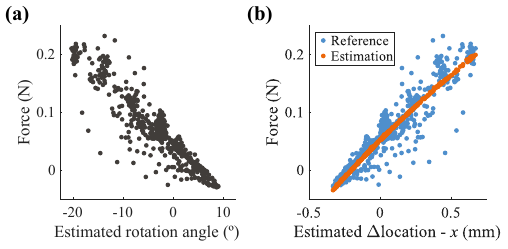}
    \caption{Characterization results.
    (a) Linear relationship between the fin rotational angle estimated from the magnet location and the force applied by the flow.
    (b) Mapping results of applied forces and estimated magnet location change using the second-order fitting method.
    }
    \label{fig:FinSensor_03}
\end{figure}
The location of the magnet, estimated using the model presented in Section~\ref{sec:FinSensor_Model}, is mapped to the force applied to the fin by the flow through fitting and ML approaches, similar to the foot contact sensor.
In more detail, the magnetic flux with noise reduced by a low-pass filter is substituted into the three equations (\ref{eqn:magnetic_flux_density_3}) to estimate the magnet’s location ($p_{x}$ and $p_{y}$). This location enables estimation of the fin’s rotational angle, as shown in Fig.~\ref{fig:FinSensor_03}(a). In addition, using the estimated $x$- and $y$-location changes of the magnet as inputs, the mapping to force resulted in RMSE values of \SI{0.023}{\newton} for the fitting method (Fig.~\ref{fig:FinSensor_03}(b)) and \SI{0.024}{\newton} for the ML method\footnote{For further details, refer to footnote 1.}. Since the fitting method offers both higher accuracy and low computational cost, it is selected.

\section{Robot platform}

To equip the previously developed salamander robot, Polymander \cite{geversInvestigatingEffectMorphology2025}, with the force sensors, we developed new sealing mechanisms and added a dedicated serial bus line to control the sensors.

\subsection{Mechanical Design}

\subsubsection{Overall structure}

The Polymander II has an overall dimension of 1.11 L $\times$ 0.39 W $\times$ 0.13 H m and weighs \SI{2.71}{\kilo\gram}. The skeleton of Polymander II contains 16 actuated DoFs (Fig.~\ref{fig:fig_overall}): 8 for lateral bending of the spine (1 for head, 4 for trunk, and 3 for tail) and 2 for each leg (fore-aft and dorsoventral bending). Each DoF is actuated by a Dynamixel XM430 servo motor (ROBOTIS, Korea). 
Custom connectors between the servo motors were printed with carbon fiber reinforced Nylon on a Markforged 2 3-D printer. All the other custom connectors were printed using Polymaker Polymax Polylactic acid (PLA) on a Bambu Lab P1S 3-D printer. 

Different from~\cite {geversInvestigatingEffectMorphology2025}, a flexible passive tail is clamped to the posterior end of the actuated tail outside the suit to increase swimming performance without adding more motors~\cite{leftwichWakeStructuresSwimming2012, karakasiliotis2016CineradiographyBiorobotsApproach, anastasiadisEellikeRobotSwims2024}. The flexible tail was 3-D printed with Recreus Thermoplastic Polyurethane (TPU) 70A on a Creality Ender-3 S1 Plus 3-D printer~\cite{yasui2025MultisensoryFeedbackMakes}. The tail is \SI{0.28}{\meter} long, \SI{0.04}{\meter} high, and weighs \SI{0.1}{\kilo\gram}.
In addition, a passive pitch joint connects the actuated tail skeleton and the hind girdle, and another connects the passive tail and the actuated tail. This improves walking balance by allowing a smooth, 3-D printed PLA component contacting the ground to support part of the tail’s weight.

To install the robot skeleton inside the dry suit (\ref{section_sealing}) and to allow easier reconfiguration into different morphologies \cite{geversInvestigatingEffectMorphology2025}, we divide the skeleton into 5 modules: head, trunk, tail, and front and hind legs. The components inside each module were preassembled. To connect the modules inside the suit (see the supplementary video for the procedure), two types of compact quick connectors were designed for connections that require passive rotation (between the tail and the hind girdle) and for fixed connections (between the legs, the head, and the trunk).

\subsubsection{Waterproofing} \label{section_sealing}
\begin{figure}
    \centering
    \includegraphics[width=1.0\linewidth]{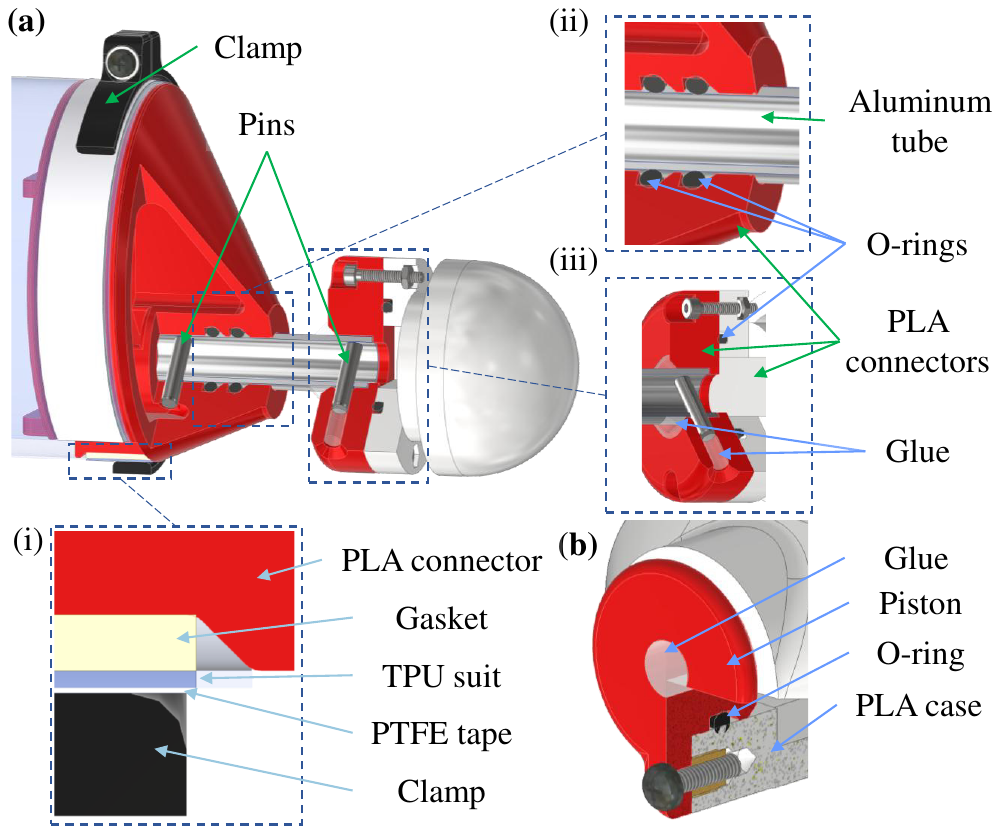}
    \caption{Sealing mechanisms.
    (a) Sealing structure for the foot. (i) Hose clamp sealing for TPU suit. (ii) Rod sealing for lower leg tube. (iii) Axial sealing for foot contact sensors.
    (b) Piston sealing for sensor modules.
    }
    \label{fig:fig_sealing}
\end{figure}

The robot is waterproofed using a transparent TPU suit~\cite{liu2025ArtificialEmbodiedCircuits} and custom 3-D printed mechanisms to meet various requirements (Fig.~\ref{fig:fig_sealing}). This sealing solution enables visual inspection of the skeleton and easy disassembly for maintenance, while providing air-tightness for leak detection and reducing deformable buoyant volumes.

We used a custom inflatable TPU structure (Kunshan Pinhong EPT, China), which was manufactured by high-frequency welding of TPU tubes and commonly used for inflatable boats, as the dry suit for the skeleton. 
With a thickness of \SI{0.3}{\milli\meter}, the suit allows each leg to bend $\pm$~\ang{73} along the yaw axis or \ang{-72} to \ang{105} along the pitch axis. An air outlet was welded to the suit (Fig.~\ref{fig:fig_overall}(a), above the middle of the trunk region) for inflation to check leakage points or deflation to reduce the volume. 
Six openings were made near the head, the tail, and four feet to install the robot skeleton. To seal the openings, we clamped the TPU suit around 3-D printed connectors with 3-D printed hose clamps (Fig.~\ref{fig:fig_sealing}(a)(i))\cite{liu2025ArtificialEmbodiedCircuits,baines2022MultienvironmentRoboticTransitions}. We used compliant rubber gaskets and PTFE tape to fill the gaps between the TPU suit and the plastic parts.

To connect each foot contact sensor with the leg inside the dry suit, we used an aluminum tube as the lower leg link (Fig.~\ref{fig:fig_sealing}(a)(ii)). The tube was fixed to the PLA connectors with dowel pins and sealed radially using two nitrile butadiene rubber (NBR) 70A O-rings.
To seal the Hall-effect sensors within a limited axial space, we 
pushed the PLA foot base with an NBR 70A O-ring axially against 
a 3-D printed PLA adapter. 
For connectors that need to be more frequently disconnected and bear little mechanical stress, such as those connecting the sensor modules (Section~\ref{section_electronics}), we used piston-style sealing with an NBR 70A O-ring (Fig.~\ref{fig:fig_sealing}(b)).
We used rapid adhesives and silicone glue to seal permanent joints, such as wires going through sealing pistons and PLA adapters on the aluminum tubes.

\subsection{Electronics and firmware} \label{section_electronics}
\begin{figure}
    \centering
    \includegraphics[width=1.0\linewidth]{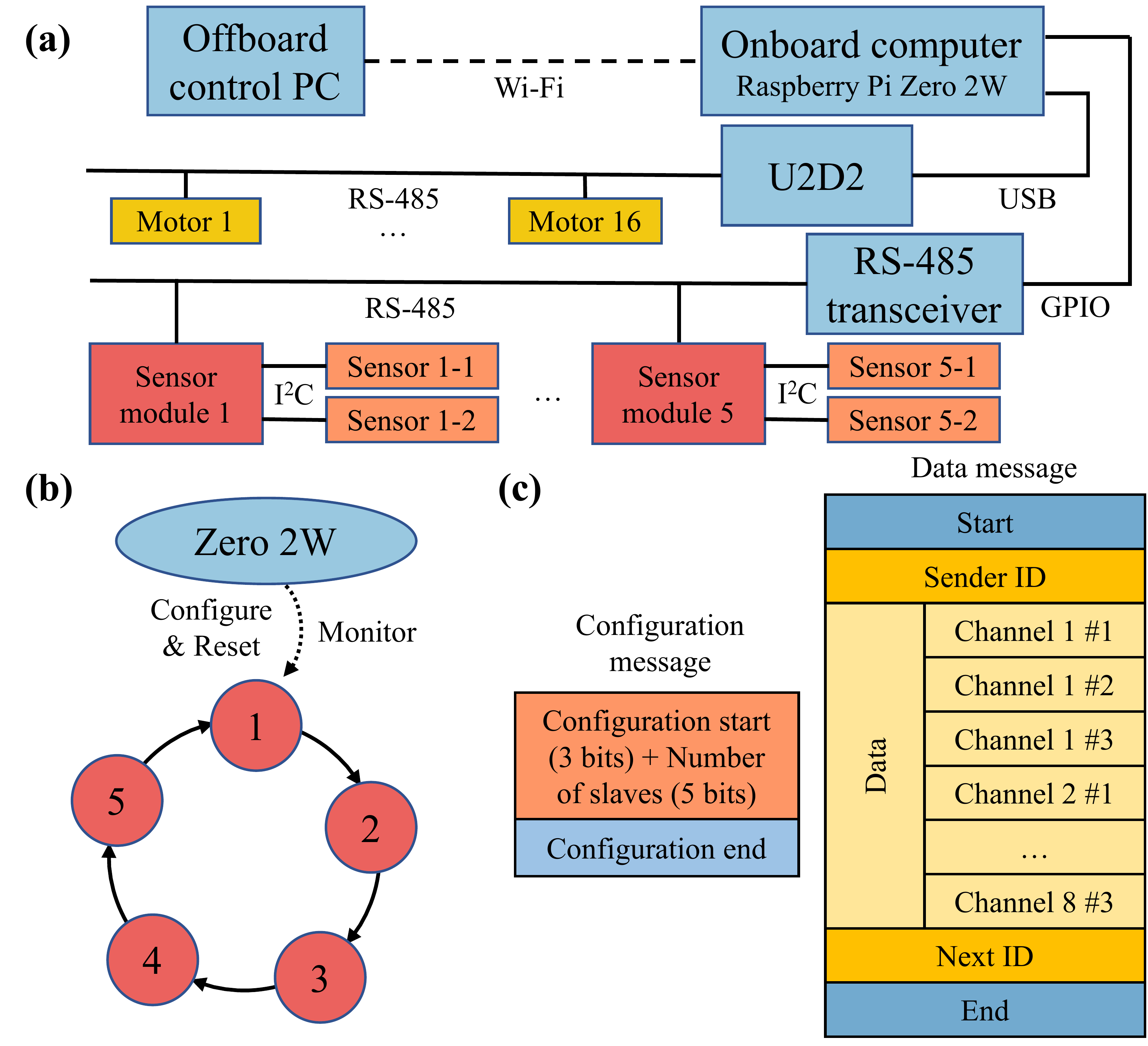}
    \caption{Electronics and firmware.
    (a) System architecture.
    (b) Communication Protocol.
    (c) Data frames.
    }
    \label{fig:fig_firmware}
\end{figure}

\subsubsection{Overall architecture}
The electronics system of the Polymander II (Fig.~\ref{fig:fig_firmware}(a)) contains an off-board control PC for user operations and future heavy computation tasks such as computing spiking neural network controllers. One onboard Raspberry Pi Zero 2W is responsible for running low-level controllers and communication with the peripherals, such as the servo motors and the sensor modules. Both computers connect to the same Wi-Fi network to allow remote communication via SSH or ROS2. 

\subsubsection{Communication}
To interface with multiple peripherals at high speeds
and allow easy addition or removal of peripherals, the onboard computer manages two bus lines, one for the servo motors and the other for the sensors. 
Both bus lines allow daisy chaining of devices, making wire management much easier.

The onboard computer connects to the motor bus line using a USB to RS-485 dongle (ROBOTIS U2D2). 
This allows us to write to each motor within \SI{2}{\micro\second} and read from one motor within \SI{0.3}{\milli\second}. 

We developed custom circuits to connect the TMAG 5273 Hall-effect sensors to the sensor bus line via I$^{2}$C lines. The circuit also contains a Seeed Studio XIAO RP2350 microcontroller
and connectors to daisy chain other sensor modules on the bus line. We also used an SP3485 RS-485 transceiver to convert the TTL-level UART serial signals from the onboard computer into the sensor bus line.

To increase the speed of communicating with a large number of modules, we developed firmware with the following attempts:
\begin{itemize}
    \item We utilize both cores of the RP2350 controller, one for reading the magnetic flux along 3 axes and temperature from the Hall-effect sensors, and the other for communicating with the bus line.
    
    \item We use a masterless token ring protocol to exchange data between the onboard computer and sensor modules (Fig.~\ref{fig:fig_firmware}(b)). In most commercial solutions, such as Modbus RTU, high communication latency is induced by each module waiting for requests from the onboard computer. Instead, here each module transmits data immediately after the previous module finishes its transmission. The onboard computer keeps reading the bus line and only sends configuration messages to start or reset the data exchange.
    
    \item We use a custom data frame definition (Fig.~\ref{fig:fig_firmware}(c)) to reduce the number of bytes needed to transmit the information. 
\end{itemize}

\subsubsection{ROS2}
We use the Robot Operating System 2 (ROS2) to program the controller~\cite{kronauerLatencyAnalysisROS22021}, containing three main nodes that communicate with each other: (1) a central controller node that computes goal joint angles at \SI{100}{\hertz}; (2) a motor control node that sends goal angles to servo motors and reads motor states, including goal positions received, present positions, voltages, and currents, at \SI{200}{\hertz}~\cite{km-robota2025KMR_dxlSourceMain, 2025ROBOTISEManual}; and (3) a sensor communication node that interprets sensor readings as soon as it receives them.

\subsubsection{Docker}
The Zero 2W computer is compact at the cost of limited computational power. To enable easier development of the controller, we use Docker to (1) encapsulate the controller with the entire software environment into a single container image that can be built and executed with simple commands, and (2) cross-compile the image on a powerful PC that can be deployed quickly on the Zero 2W.

\subsection{Central pattern generator controller for walking and swimming}

The swimming and walking of the robot was controlled by a CPG controller~\cite{ijspeert2007swimming}.
Each oscillator $i$ is described by two differential equations for its phase $\phi_i$ and amplitude $r_i$, and an equation for its final activity output $x_i$:
\begin{equation}
    \begin{aligned}
        &\dot{\phi}_i = \omega_i + \sum_j \text{w}_{ij} \sin(\phi_j - \phi_i - b_{ij})\text{,}\\  
        &\dot{r}_i = a_i (R_i - r_i)\text{,}\\
        &\text{and}~x_i = r_i (1+cos(\phi_i))
    \end{aligned}
    \label{eq:oscillator}
\end{equation}
where $\phi_i$ and $r_i$ are the phase and amplitude of oscillator $i$, $\omega_i$ is the intrinsic angular frequency, $\text{w}_{ij}$ and $b_{ij}$ are the coupling strength and phase bias between oscillators $i$ and $j$, $a_i$ is a time constant and $R_i$ is the target amplitude of the oscillator. Both $\omega_i$ and $R_i$ are modulated by descending drive signals.

Each active joint in the robot is controlled by a pair of oscillators. The target position is obtained by differentiating the output activities of the oscillator pair. The network's connectivity was set using the same architecture as~\cite{ijspeert2007swimming}.

By adjusting the descending drive signals, the same network generates distinct locomotor patterns. Low descending drive amplitudes activate both limb and axial oscillators, producing a standing-wave pattern of trunk undulations coupled to stepping, corresponding to walking. As the drive increases, the limb oscillators saturate and effectively decouple from the network, allowing the axial oscillators to synchronize into a traveling wave pattern characteristic of swimming.

The controller was developed and validated in FARMS (Framework for Animal and Robot Modeling and Simulation)~\cite{arreguit2025farms}, which allowed rapid prototyping and testing of oscillator network configurations. Simulations were performed in MuJoCo~\cite{todorov2012mujoco} to capture the dynamics of the articulated body and its interactions in ground and water environments. 



\section{Experimental Setup}

To validate that the foot contact and flow sensors detect the environmental interactions,
we recorded sensor readings and videos when the robot was walking on the lab floor or swimming in a swimming pool (6 m long, 2 m wide, and 0.3 m deep).
During walking experiments, the displacement of markers attached to the robot head was tracked by a Vicon motion capture system.
To ensure robust transfer from simulation to hardware, the robot was controlled in open loops by replaying CPG-generated kinematics from simulation. 

To demonstrate the potential of multimodal sensory feedback control, we also implemented a simple controller that triggers the transition from walking to swimming using contact force feedback, which was previously performed manually \cite{ijspeert2007swimming,crespiSalamandraRoboticaII2013}. The descending drive signal was initialized for walking and switched to the swimming value when the sum of the $x$-axis forces on four feet fell below 7 N. Sensor readings were first validated under open-loop control using replayed kinematics. The feedback controller was then executed on the onboard computer, while the operator controlled turning via a gamepad to avoid collisions with the pool walls. The sensory data rate was limited to 100 Hz and the control loop to 50 Hz to reduce computational load. More advanced control strategies leveraging these sensors will be explored in future work.

\section{Experimental Results}


\subsection{Walking}

\begin{figure}
    \centering
    \includegraphics[width=1.0\linewidth]{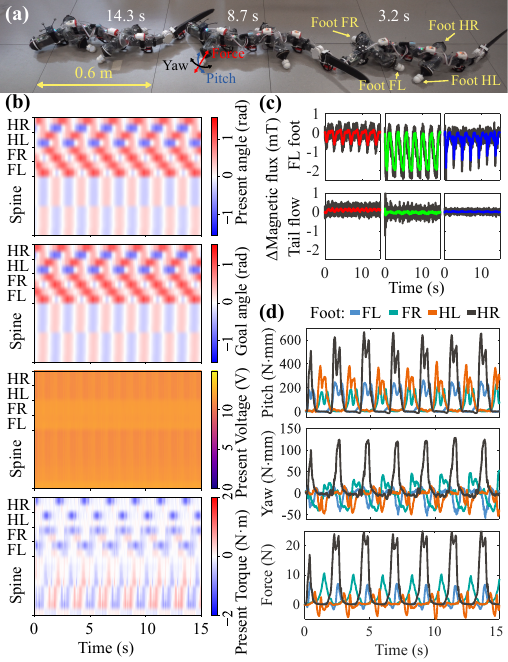}
    \caption{Walking with an open-loop CPG controller.
    (a) Snapshots of a walking sequence.
    (b) Readings of servo motors in the spine, front left (FL), front right (FR), hind left (HL), and hind right (HR) legs.
    (c) Raw (gray) and smoothed (colored) readings of Hall-effect sensors.
    (d) Interpreted foot contact force and torques.
    }
    \label{fig:fig_walking}
\end{figure}

With a gait frequency of \SI{0.47}{\hertz}, Polymander II achieved a stride length of 0.341~$\pm$~\SI{0.012}{\meter} or 0.641~$\pm$~0.023 snout-vent length, measured from the anterior end to the center of the hind girdle (Fig. \ref{fig:fig_walking}(a)), in four trials. Despite the additional weight and volume taken by the Hall-effect sensors, the walking performance was comparable to salamanders and previous salamander robots after scaling~\cite{crespiSalamandraRoboticaII2013,karakasiliotis2016CineradiographyBiorobotsApproach}.

Polymander II read all the force sensors at 589.80~$\pm$~\SI{0.08}{\hertz} and read the 16 motors at 99.9~$\pm$~\SI{0.20}{\hertz}. The motor torques (Fig.~\ref{fig:fig_walking}(b)), estimated by the voltage and current readings, peaked at \SI{1.33}{\newton\meter}, with still some margins from the motors' maximum ratings~\cite{2025ROBOTISEManual} for more demanding tasks. The low-pass filters effectively filtered out the noises without causing significant delays (Fig.~\ref{fig:fig_walking}(c)). 
The flow sensor readings remained around zero, despite their sensitivity to small force readings, which shows the small effects of inertial forces when they are moving in the air and the effectiveness of the cages protecting them. 
The foot contact sensor readings showed repeatable periodic patterns. The normal ground reaction forces, contributing mainly to the pitch torque and the force readings, were in phase between diagonal feet and out of phase between ipsilateral feet (Fig. \ref{fig:fig_walking}(d)), in line with the gait cycles recorded. The yaw torques, contributed mainly by the friction, were smaller than the pitch torques, because of the low friction coefficient between the PLA foot skin and the smooth lab floor.
However, the magnitudes appeared unbalanced between different feet. This results from nonlinear behavior under large deformations, which leads to increased errors under high three-dimensional loads. There is a trade-off between accuracy under large forces and sensitivity to small loads.

\subsection{Swimming}
\begin{figure}
    \centering
    \includegraphics[width=1.0\linewidth]{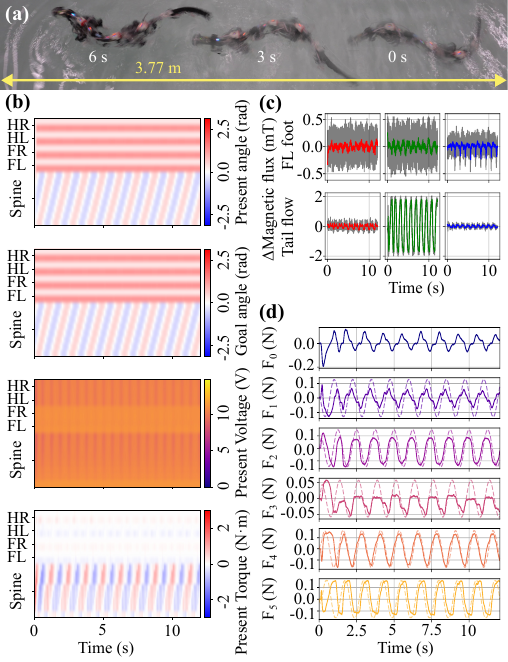}
    \caption{Swimming with an open-loop CPG controller.
    (a) Snapshots of a swimming sequence.
    (b) Readings of servo motors.
    (c) Raw (gray) and smoothed (colored) readings of Hall-effect sensors.
    (d) Interpreted hydrodynamic force readings (solid) compared to scaled angles of joints anterior to corresponding links (dashed).
    }
    \label{fig:fig_swimming}
\end{figure}

With a gait frequency of \SI{0.78}{\hertz} and body wave amplitude of \ang{29}, the robot traversed the camera field of view (Fig. \ref{fig:fig_swimming}(a); \SI{3.77}{\meter} wide) in 12.23~$\pm$~\SI{0.60}{\second} in three trials, reaching a speed of 0.28~$\pm$~0.01 body length per second. The agreement of swimming speed with previous robots and salamanders after scaling~\cite{crespiSalamandraRoboticaII2013,karakasiliotis2016CineradiographyBiorobotsApproach} demonstrates that the sensor structures did not compromise swimming performance.

Polymander II read the 16 servo motors at 99.88~$\pm$~\SI{0.17}{\hertz} and read the 4 contact sensors and 6 flow sensors at 593.80~$\pm$~\SI{2.53}{\hertz}.
The foot contact sensor readings stayed around zero (Fig.~\ref{fig:fig_swimming}(c)) compared to walking, suggesting minimal effects of the hydrodynamic forces on them, which can be utilized to detect transitions between aquatic and terrestrial environments. In contrast, the flow sensors were sensitive enough to measure the hydrodynamic forces (Fig.~\ref{fig:fig_swimming}(d)). As the spine propagated a traveling wave down the body, the flow sensor readings were also propagated from head to tail. Each sensor reading was synchronized with the angle of the joint anterior to the local link (Fig.~\ref{fig:fig_swimming}(d), solid versus dashed), with minor differences in phase lags resulting from different installation locations. This is similar to ~\cite{thandiackalEmergenceRobustSelforganized2021} (angles flipped to match the definitions), suggesting potential usage of flow sensory feedback to supplement joint angle feedback for additional robustness. The amplitudes of the flow sensor readings were mostly consistent along the body, except for the two under the girdles ($F_1$ and $F_3$), likely because the legs affected the local flow.

\subsection{Transition from walking to swimming}

Both the foot contact force sensors and the flow sensors showed distinct reading patterns before and after the transition (see supplementary video). 
Despite the lower control frequency, contact force readings successfully triggered the transition from walking to swimming when the robot stepped into water and floated on the surface (see supplementary video).

\section{Discussion and Conclusions}
In this work, we developed contact and flow sensors based on Hall-effect sensors and equipped them to an amphibious, salamander-inspired legged robot. The sensors sensed environmental interaction, including ground contact forces and hydrodynamic forces, at high speeds. The robot demonstrated fast walking and swimming and automatically transitioned between them using contact force feedback. This provides a solid foundation for future verification of feedback controllers for amphibious robots and holds promise in the fields of both robotics and neuromechanics. 

Future work on the hardware design includes increasing the robustness of the sensors, adding tactile sensors to more locations, and developing more integrated mechatronic structures to increase the modularity. More effective calibration pipelines remain to be developed to reduce the nonlinear errors. Furthermore, feedback controllers that can fully utilize the sensory information need to be investigated. Both the controllers and hardware should be more systematically tested across different speed regimes and environmental conditions to better understand how sensory feedback can contribute to multimodal locomotion.

\addtolength{\textheight}{0cm}   

\section*{Acknowledgment}
The authors would like to thank Alessandro Crespi, François Longchamp, Xiangxiao Liu, and Kamilo Melo for advice on hardware design.

\bibliographystyle{IEEEtran}
\bibliography{IEEEabrv,bibtex}

\end{document}